\documentclass[10pt, conference]{IEEEtran}
\IEEEoverridecommandlockouts
\usepackage{cite}
\usepackage{amsmath,amssymb,amsfonts}
\usepackage{algorithmic}
\usepackage{graphicx}
\usepackage{textcomp}
\usepackage{xcolor}
\usepackage{url}
\usepackage{hyperref}
\usepackage{array}
\usepackage{caption}
\usepackage{todonotes}
\usepackage{floatrow}
\usepackage{subfig}

\usepackage{algorithm}

\newfloatcommand{capbtabbox}{table}[][\FBwidth]

\usepackage{blindtext}

\renewcommand{\baselinestretch}{0.94}
\begin{document}

\title{Budget-based real-time Executor for Micro-ROS\\
\thanks{This  work  has  been  supported by the EU-funded project OFERA (micro-ROS) under Grant No. 780785. 
Many thanks go to eProsima for support on the setup of the Olimex board and NuttX.}
}

\author{\IEEEauthorblockN{1\textsuperscript{st} Jan Staschulat}
\IEEEauthorblockA{\textit{Robert Bosch GmbH} \\
Stuttgart, Germany \\
jan.staschulat@de.bosch.com}
\and
\IEEEauthorblockN{2\textsuperscript{nd} Ralph Lange}
\IEEEauthorblockA{\textit{Robert Bosch GmbH} \\
Stuttgart, Germany \\
ralph.lange@de.bosch.com}
\and
\IEEEauthorblockN{3\textsuperscript{rd} Dakshina Narahari Dasari}
\IEEEauthorblockA{\textit{Robert Bosch GmbH} \\
Stuttgart, Germany \\
dakshina.dasari@de.bosch.com}
}

\maketitle

\begin{abstract}
The Robot Operating System (ROS) is a popular robotics middleware framework. In the last years, it underwent 
a redesign and reimplementation under the name ROS~2. It now features QoS-configurable communication and a 
flexible layered architecture. Micro-ROS is a variant developed specifically for resource-constrained 
microcontrollers (MCU). Such MCUs are commonly used in robotics for sensors and actuators, for time-critical control 
functions, and for safety. While the execution management of ROS 2 has been addressed by an Executor concept, its lack 
of real-time capabilities make it unsuitable for industrial use. Neither defining an execution order nor the 
assignment of scheduling parameters to tasks is possible, despite the fact that advanced real-time scheduling algorithms are 
well-known and available in modern RTOS's. For example, the NuttX RTOS supports a variant of the reservation-based scheduling which 
is very attractive for industrial applications: It allows to assign execution time budgets to software components 
so that a system integrator can thereby guarantee the real-time requirements of the entire system. This paper presents for the 
first time a ROS~2 Executor design which enables the real-time scheduling capabilities of the operating system. In particular, we 
successfully demonstrate the budget-based scheduling of the NuttX RTOS with a micro-ROS application on an 
STM32 microcontroller. 
\end{abstract}


\section{Introduction}
\noindent The Robot Operating System (ROS) is a popular robotics middleware framework with roots in academic research.
In the last years, it underwent a redesign and reimplementation under the name ROS 2~\cite{ROS2} to address the needs of industry.
It now features QoS-configurable communication by adopting the Data Distribution Service (DDS) standard, a flexible layered architecture, support for macOS and POSIX-compliant OSes in general as well as Windows, and built-in mechanisms for multi-robot systems. Micro-ROS~\cite{microROS} is a variant developed specifically for resource-constrained microcontrollers (MCU) that feature few tens or hundreds kilobytes of RAM only. Such MCUs are used commonly in robotics for sensors and actuators, for time-critical control functions, for power efficiency, and for safety.
This applies in particular for industrial applications.
The goals of micro-ROS are two-fold: First, it integrates MCUs with ROS 2 so that the software on an MCU can be accessed from the main processor by the standard ROS 2 communication mechanisms and tools.
Second, it brings major concepts and interfaces of the ROS 2 Client Library to MCUs so that software can be ported easily.
The communication in micro-ROS is based on the new DDS-XRCE standard, which integrates seamlessly with DDS by an agent running on the main processor.

Real-time support is an important feature for micro-ROS, given that all robotic applications require some level of real-time performance: e.g., consider the time between detecting an obstacle by a mobile robot and reacting to it, or more generally, sense-plan-act control loops where the end-to-end latency must be guaranteed for correctness and safety.
To ensure such real-time guarantees, appropriate dispatching and scheduling support must be built into the framework.
This is neither the case in ROS 2 nor in micro-ROS, despite the availability of real-time OSes featuring formally proven and tested real-time scheduling algorithms.

There are two high-level requirements to such real-time support: First, it shall provide an easily usable interface that allows specifying the required real-time guarantees without in-depth understanding of the underlying scheduling mechanisms. This is especially important for the many SMEs in the robotics market that cannot afford dedicated real-time experts.
Second, it has to map the event-driven execution model of ROS to the scheduling mechanisms of the underlying OS.

\begin{figure}
	\centering
	\includegraphics[width=0.8\linewidth]{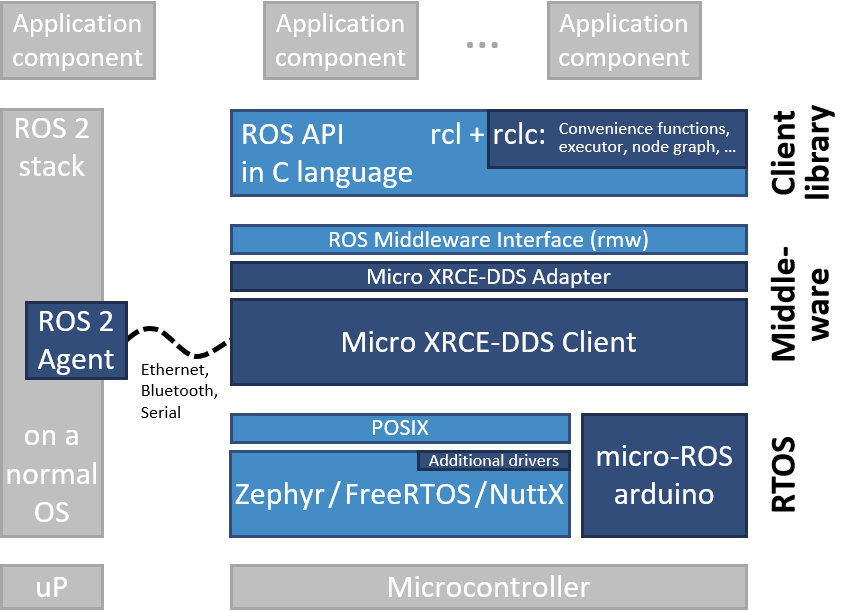}
	\caption{Architecture Stack: The dark blue components are specifically designed for Micro-ROS.}
	\label{fig:architecture}
\end{figure}

ROS 2 introduced the concept of the Executor to enable custom mappings of ROS components (called \emph{nodes}) to threads as well as to allow for application-specific extensions.
The rclc Executor of micro-ROS gives even more control by a fine-grained interface that allows registering each event (e.g., arrival of a message, timer, interrupt) individually and by separating between the detection of an event and the trigger for the corresponding event handler function.
Implementing one or even multiple time-critical functions with these Executors still requires in-depth real-time expertise.
The concept of reservation-based or \emph{budget-based scheduling} \cite{SchedulingSporadic1989} provides a much more accessible interface, which is suitable in particular for developers without such expertise.
In general, budget-based scheduling is a composable mechanism to provide quality of service guarantees to different applications.
With budget-based scheduling, every application (groups of threads or processes) is assigned a budget, which is replenished regularly.
When the application exceeds its budget, it is suspended.
With this, applications can be temporally isolated and guarantees on execution behavior can be ascertained.
New applications can be added to a running system without affecting the guaranteed performance of applications.

In this work, we investigate how to leverage budget-based scheduling for micro-ROS, by example of the sporadic scheduler offered by the NuttX OS~\cite{Nuttx}.
The most important question to answer is, how to design an executor which not only manages how the incoming events are dispatched to callbacks but also how these invocations are tied with the underlying scheduling primitives.
In detail, our contributions are:
\noindent
\begin{enumerate} 
	\item Redesign of the micro-ROS rclc Executor for real-time-capability.
	\item Concept for the use of budget-based sporadic scheduling to enforce quality of services to different callbacks.
	\item Developer-friendly interface to specify real-time requirements.
	\item Validation of the proposed design by a prototypical implementation for micro-ROS on embedded board with an STM32 Cortex-M4 MCU.
\end{enumerate}
\section{Background}

\subsection{Execution Management in ROS}
\noindent ROS applications are organized around self-contained functional units called \emph{nodes}.
Communication between the nodes is facilitated via the publish-subscribe paradigm, wherein nodes publish messages onto specific topics, these messages are broadcasted, and other nodes, that subscribed to the relevant topic, receive these messages.
Similarly, ROS also features the client-server paradigm.
Nodes react to the incoming messages by invoking callbacks to process each message.
This invocation is performed by the \emph{Executor}.
In detail, the Executor coordinates the execution of callbacks issued by the participating nodes by checking the incoming messages from the DDS queue and dispatching them to the underlying threads for execution.
Currently, the dispatching mechanism is very basic: the Executor looks up wait queues, which notifies it of any pending messages in the DDS queue.
If there are pending messages, the Executor simply executes them one after the other, which is also called round-robin.
In addition, the Executor also checks for events from application-level timers, which are always processed before messages \cite{Casini_et_al_2019_Executor_Analysis}.
There is no further notion of prioritization or categorization of the incoming callback calls.
Moreover, the Executor in its current form, does not leverage the real-time capabilities of the underlying OS scheduler to have finer control on the order of executions.
The overall implication of this behavior is that time-critical callbacks could suffer possible deadline misses and degraded performance since they are serviced later than non-critical callbacks.
Additionally, due to the round-robin mechanism and the resulting dependencies it causes on the execution time, it is difficult to determine usable bounds on the worst-case latency that each callback execution may incur. 

\subsection{Rclc Executor}
\noindent Execution management in micro-ROS follows the same principles as in standard ROS 2: The DDS-XRCE middleware is integrated via the same \emph{ROS Middleware Interface} (rmw), 
client library concepts such as nodes, subscriptions and callbacks are the same -- and either even based on the same \emph{ROS Support Client Library} (rcl) or implemented by a 
supplementary package named \emph{rclc}. Together, rcl and rclc provide a feature-complete client library in the C programming language. However, micro-ROS comes with its own Executor, 
provided by the rclc package \cite{JanEmsoft}.

%

\subsection{NuttX and its Sporadic Scheduler}
\noindent NuttX~\cite{Nuttx} is a POSIX- and ANSI-compliant open-source operating system with a small footprint, catered towards tiny to small deeply embedded environments.
With a full preemptible tickless kernel, support for fixed priority, round robin, the POSIX defined sporadic scheduling, and priority inheritance, NuttX is real-time in nature. 

The POSIX sporadic scheduling policy offered by NuttX is a variant of the Sporadic Server~\cite{SchedulingSporadic1989} and is used to enforce an upper limit on the execution time of a thread within a given period of time. 
Each thread is associated with a budget, a replenishment period, a normal priority, a low priority and the maximum number of replenishments.
The budget refers to the amount of time a thread is allowed to execute at its normal priority before being dropped to its low priority.
The replenishment period corresponds to the interval during which the budget can be consumed and is also used to compute the next replenishment time of a sporadicly scheduled thread.
The maximum number of replenishments limits the number of replenishment operations that can take place, thereby bounding the amount of system overhead consumed by the sporadic scheduling policy.
If a thread $\tau_i$ with period $T_i$ and budget $B_i$, consumes a budget of $b$ starting from time $t_x$, then $\tau_i$ receives a  replenished budget of $b$ at time $t_x + T_i$.
As soon as $\tau_i$ consumes $B_i$ units of budget in a period of $T_i$ units of time, its priority drops and it stops executing at its normal priority.

As in FIFO scheduling, a thread using sporadic scheduling continues executing until it blocks or is preempted by a higher-priority thread.
Thus sporadic scheduling does not guarantee that every thread will receive its guaranteed budget, it only ensures that threads do not consume more than their assigned budgets (in the presence of other ready threads).
The provision of an additional background or low priority helps by avoiding hard-budgeting which can be inefficient and lets the threads use the processor time efficiently, albeit at a lower priority.

\section{Budget-based real-time rclc-Executor}
\noindent We develop our real-time Executor on the NuttX RTOS running on an STM32F407 microcontroller. 

\begin{figure}
	\centering
	\includegraphics[width=1\linewidth]{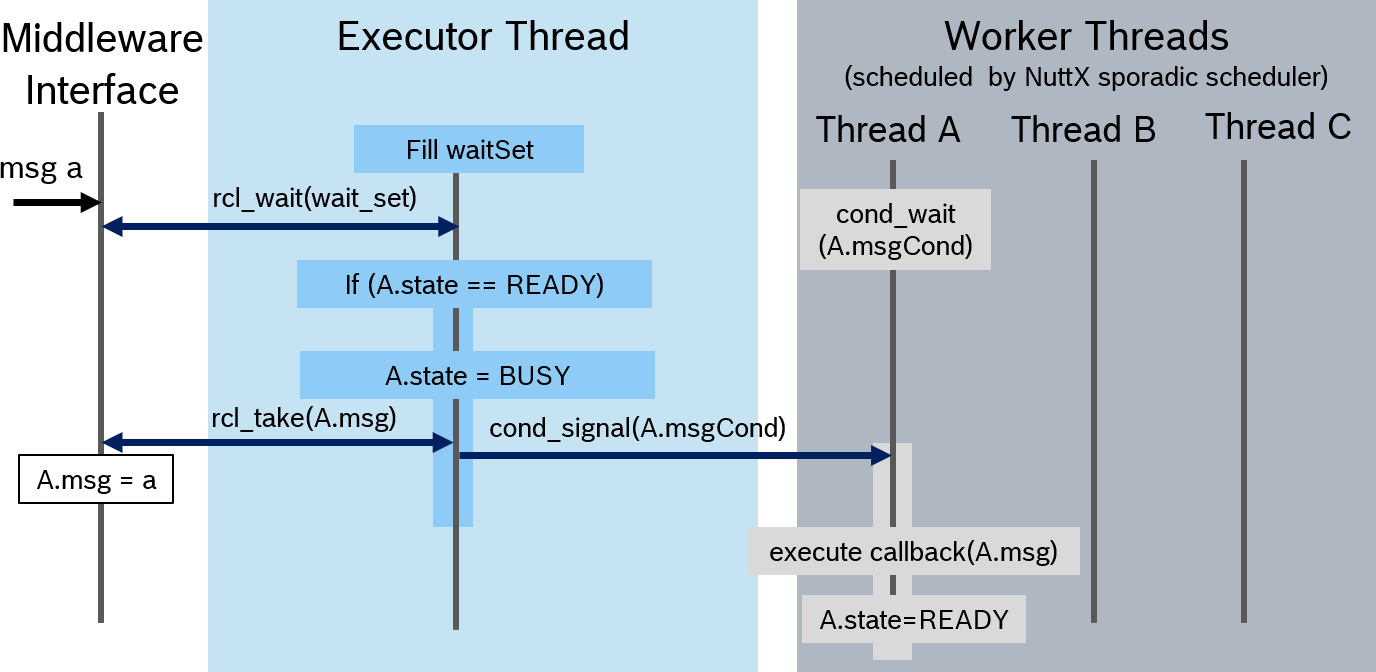}
	\caption{Interaction diagram of real-time rclc Executor.}
	\label{fig:executordesign}
\end{figure}

Figure~\ref{fig:executordesign} shows the overall architecture: one executor thread communicates with the middleware and multiple worker thread are dedicated for executing 
subscription callbacks. The implementation of the executor thread is shown in Algorithm~\ref{alg:executor-thread}. After spawning the worker threads with the user-defined scheduling policies,
it waits for new messages in the DDS-queue. However, it fetches a message only from the DDS-queue if the 
corresponding worker thread is also ready. If so, it signals the worker thread to process the message. 
%


\small
\begin{algorithm}
    \caption{Executor thread}
    \label{alg:executor-thread}
	\small
    \begin{algorithmic}[1]
		\STATE initialize $waitSet$
		\STATE allocate and initialize array $workerThread[]$
		\FOR {$s_i \in Subscriptions$}
			\STATE {\tt pthread\_create($s_i.wThread, s_i.schedP,$ \newline
			    $workerThreadRun, workerThread[i]$)}
        \ENDFOR
			\WHILE{ $rcl\_context\_is\_valid()$ }
		   \STATE fill $waitSet$ with subscriptions
		   \STATE {\tt rcl\_wait($waitSet$, $timeout$)}
		   \FOR{ all subscriptions $s_i \in waitSet$ with new data}	   
				\IF{$workerThread[i].state == READY$}
					\STATE {\tt rcl\_take($s_i$, $workerThread[i].msg$)}
					\STATE $ workerThread[i].state  \leftarrow BUSY$
					\STATE {\tt cond\_signal($workerThread[i].msgCond$)}
				\ENDIF
		   \ENDFOR
		\ENDWHILE
    \end{algorithmic}
\end{algorithm}
\normalsize

The worker thread is shown in Algorithm~\ref{alg:worker-thread}.
It waits for a new message by waiting for the condition variable $msgCond$.
When a new ROS message is available, then the worker thread executes the corresonding callback. 
The scheduling parameters of the worker thread were already configured by the executor thread (line 4).
\begin{algorithm}
    \caption{workerThreadRun(void * param)}
    \label{alg:worker-thread}
	\small
    \begin{algorithmic}[1]
		\STATE $p \leftarrow (cast) param$
		\WHILE{ 1 }
		   \STATE {\tt cond\_wait(p.msgCond, p.workerMutex)}
		   \STATE $p.callback(p.msg$)
		   \STATE $p.state \leftarrow READY$
		\ENDWHILE
	\end{algorithmic}
\end{algorithm}

For simplicity, the description of the mutex to protect accesses to shared data structures between the 
worker thread and executor thread has been omitted. 

The current version of the micro-ROS Middleware (i.e., of DDS-XRCE) is single-threaded, that is
all accesses to ROS middleware must be protected by a mutex, e.g. $mw\_mutex$. This includes 
{\tt rcl\_wait, rcl\_take} and {\tt rcl\_publish}. 
Therefore we define a wrapper {\tt rclc\_executor\_publish()}, which takes care of the lock, and shall be used in 
the user callback functions.


The user interface is shown in Algorithm~\ref{alg:user-interface}. First the scheduling parameters, especially the RTOS-specific
{\tt sched\_param} must be defined. Secondly, these parameters are passed with the subscription, when it is added to the executor.
Note that the executor design is compliant with any POSIX-like operating system that provides similar interfaces.  
In this paper we present a general approach to define the budget-based scheduling, with NuttX RTOS as an example. It is available in the rclc package~\cite{GithubRclcExecutor}.


\begin{algorithm}
	\caption{User Interface}
    \label{alg:user-interface}
{\small
\begin{verbatim}
	rclc_executor_sched_parameter_t sp;
	sp.policy = SCHED_SPORADIC;
	struct sched_param p;
	p.sched_priority               = 60;
	p.sched_ss_low_priority        = 6;
	p.sched_ss_repl_period.tv_sec  = 0;
	p.sched_ss_repl_period.tv_nsec = 100000000;
	p.sched_ss_init_budget.tv_sec  = 0;
	p.sched_ss_init_budget.tv_nsec = 30000000;
	p.sched_ss_max_repl       = 100;
	sp.param = p;
	rclc_executor_add_subscription_sched(&exe, \
	  &sub1, &msg1, &cb, ON_NEW_DATA, &sp));
	rclc_executor_start_multi_threading
	_for_nuttx(&exe));
\end{verbatim}
}
\end{algorithm}

\section{Experimental Setup}
\begin{figure} [h!]
 	\centering
 	\includegraphics[width=1.0\linewidth]{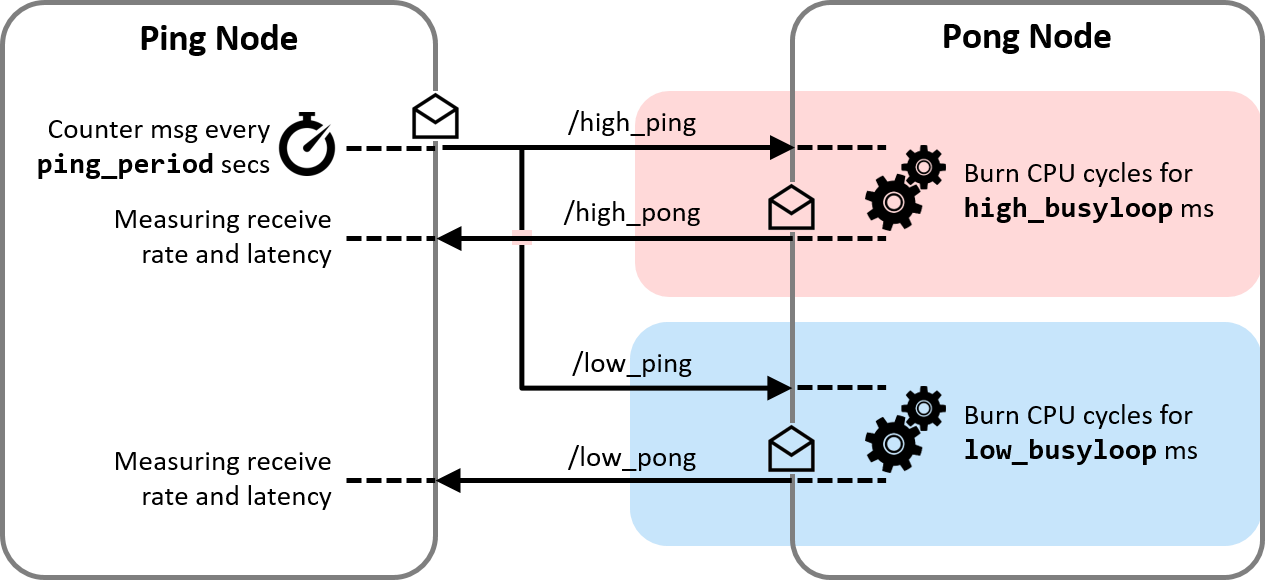}
 	\caption{TestBench Setup.}
 	\label{fig:pingpongarchitecture}
 \end{figure}

\noindent As a proof of concept, we implemented a simple ping-pong use-case as depicted in Figure~\ref{fig:pingpongarchitecture}. The Ping node runs on the Linux host while the Pong node runs on the NuttX RTOS on a microcontroller board by Olimex, hosting an STM32F407 microcontroller with $196\,\mathrm{kB}$ of RAM. The Ping node can be configured to send messages at a configured rate. The Pong node takes these 
ping messages, carries out some processing for a configurable time and then replies back with a pong message. We simulate a scenario where the two nodes exchange 
high priority real-time (HPRT) and low priority best-effort (LPBE) messages simultaneously with each other. The NuttX scheduling parameters are configured for each subscription and passed to the new real-time rclc Executor. Tests with NuttX showed, that budget enforcement only works for one thread with sporadic scheduling. 
Because of this limitation, we could only configure one sporadic thread. This demo is available as open source \cite{GithubNuttXDemo}.
 
\subsection{Comparison of FIFO and Sporadic Scheduler}
\noindent In the first scenario we schedule both the  HPRT and LPBE threads, running at priority 60 and 50 respectively (60 is higher priority) on NuttX. The proposed executor runs at a higher priority (110) than these threads. 
The Ping node sends the messages every $10\,\mathrm{ms}$ and the busy loop for the high\_pong and low\_pong burn CPU cycles for $10\,\mathrm{ms}$ before sending out the response. The experiment was run for $10\,\mathrm{s}$. 

\begin{figure}
	\centering
	\includegraphics[width=1.0\linewidth]{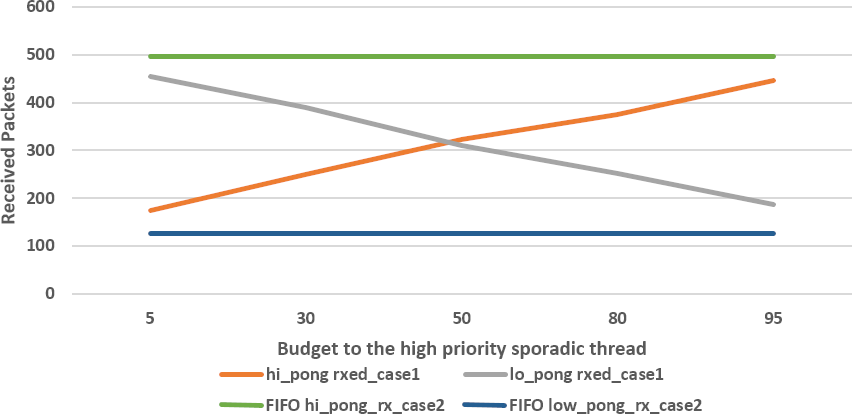}
	\caption{FIFO vs.\ sporadic scheduling with varying budgets.}
	\label{fig:exp1sporadic}
\end{figure}

\emph{Case 1:} In this scenario, we scheduled the HPRT thread with SCHED\_SPORADIC and the LPBE thread with  SCHED\_FIFO and varied the budget issued to the HPRT thread.  
Figure~\ref{fig:exp1sporadic} depicts the number of pong messages received under different budget configurations. As seen, a higher budget percentage to the sporadic 
thread ensures that more HPRT pong messages are received, confirming that the Executor is not running with the 
thread with the default FIFO and that the binding to the underlying sporadic scheduler works. The graph also shows that LPBE messages are not starved and  receive the residual budget.

\emph{Case 2:} Here we scheduled both the threads with SCHED\_FIFO and as expected, HPRTE pong messages consistently dominate the stream 
and many LPBE messages are not processed. The constant lines in Figure~\ref{fig:exp1sporadic} depict this behaviour.  However of interest is the fact that there is no direct correlation between the budget and the number for HPRT pong messages received. The reason for 
lower effective processing time available to user-defined threads can be attributed to the implementation of the NuttX stack. The interface between 
the Executor and the middleware is single threaded and other kernel artifacts (like mandatory locks) must also be accounted for and hence the 
entire CPU power cannot be made available to user-defined threads. -- Intuitively we think this is the reason why the theoretical number of messages 
are not processed, and we further plan to investigate the root cause.

\subsection{Testing work-conserving nature of the Sporadic Scheduler} 
\noindent In this experiment the HPRT ping node sends a message every $10\,\mathrm{ms}$ and the HPRT ping node has a processing time of $10\,\mathrm{ms}$. The HPRT thread is scheduled by SCHED\_SPORADIC 
and has a budget of 30\% (much lower than its utilization of 100\%), a high priority of 60, a low priority of 10 and a maximum replenishment of 100.
The LPBE ping node sends a message every $10\,\mathrm{ms}$ but the LBPE pong node does not process continuously but sleeps in the interim. The LPBE thread is scheduled by SCHED\_FIFO 
and runs at a priority of 50. Here we validate whether the sporadic thread drops to its lower priority (10) whenever its budget is exhausted, and executes at its lower
priority. In order to do so, we vary the percentage of time the LPBE thread sleeps, thereby giving a chance to the sporadic thread, when out of budget, to execute.

\begin{figure}
	\centering
	\includegraphics[width=1.0\linewidth]{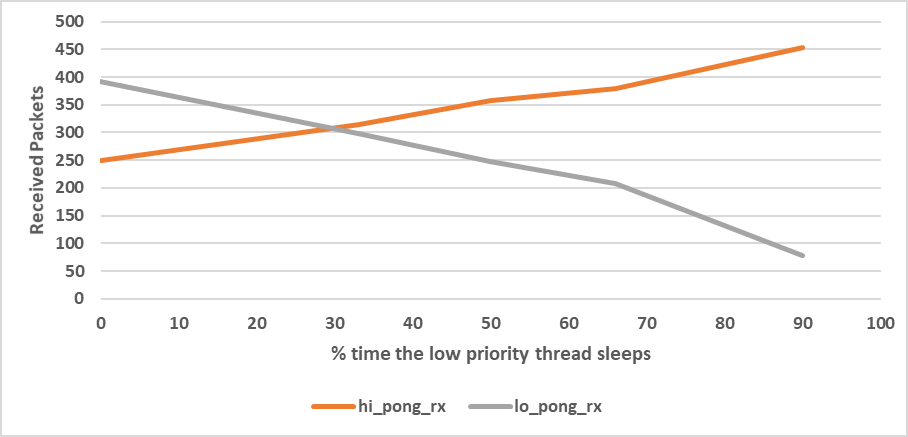}
	\caption{Work conserving nature of the sporadic scheduler.}
	\label{fig:workcon}
\end{figure}

As seen in Figure~\ref{fig:workcon}, the longer the low priority thread sleeps, the higher is the message throughput of the HPRT thread, since even if it is out of budget, it continues executing at the lower priority. This also shows that there is no hard budgeting and the scheduler is work conserving.

%
%

\section{Conclusion}
\noindent We modified the micro-ROS rclc Executor by extending the user interfaces to allow the specification of thread policies and budgets and 
by availing the real-time capabilities of the RTOS. In our experiments we have shown that the executor can realize the aforementioned 
binding to the scheduler. As future work, we plan to further refine the Executor and experiment with diverse use-cases.

\bibliographystyle{IEEEtran}
\bibliography{RTAS_21_ROS2}

\end{document}